\documentclass{article}
\usepackage{ijcai26}

\usepackage{times}
\usepackage{soul}
\usepackage{url}
\usepackage[hidelinks]{hyperref}
\usepackage[utf8]{inputenc}
\usepackage[small]{caption}
\usepackage{subcaption}
\usepackage{graphicx}
\usepackage{amsmath}
\usepackage{amssymb}
\usepackage{amsthm}
\usepackage{booktabs}
\usepackage{algorithm}
\usepackage{algorithmic}
\usepackage[most]{tcolorbox}
\usepackage{xcolor}

\definecolor{boxbg}{RGB}{240, 242, 255} 
\definecolor{boxframe}{RGB}{0, 0, 0}    
\definecolor{keywordblue}{RGB}{0, 0, 128} 
\definecolor{warningpurple}{RGB}{128, 0, 128} 

\newtcolorbox{promptbox}[1][]{
    enhanced,
    colback=boxbg,      
    colframe=boxframe,  
    boxrule=0.5pt,      
    arc=3mm,            
    auto outer arc,
    fonttitle=\bfseries\large,
    detach title,       
    before upper={\textbf{#1}\par\vspace{0.5em}}, 
    coltitle=black,
    family=serif,       
    halign=left,
    breakable,          
}
\DeclareCaptionType{promptcap}[Prompt][List of Prompts]

\title{DyACE: Dynamic Algorithm Co-evolution for Online Automated Heuristic Design with Large Language Model}

\author{
    Author Name
    \affiliations
    Affiliation
    \emails
    email@example.com
}

\author{
Guidong Lu$^1$ \and
Yiping Liu$^{2,*}$ \and
Xiangxiang Zeng$^3$
}

\begin{document}

\maketitle

\begin{abstract}
    The prevailing paradigm in Automated Heuristic Design (AHD) typically relies on the assumption that a single, fixed algorithm can effectively navigate the shifting dynamics of a combinatorial search. This static approach often proves inadequate for Perturbative Heuristics, where the optimal algorithm for escaping local optima depends heavily on the specific search phase. To address this limitation, we reformulate heuristic design as a Non-stationary Bi-level Control problem and introduce DyACE (Dynamic Algorithm Co-evolution). Distinct from standard open-loop solvers, DyACE use a Receding Horizon Control architecture to continuously co-evolve the heuristic logic alongside the solution population. A core element of this framework is the Look-Ahead Rollout Search, which queries the landscape geometry to extract Search Trajectory Features. This sensory feedback allows the Large Language Model (LLM) to function as a grounded meta-controller, prescribing phase-specific interventions tailored to the real-time search status. We validate DyACE on three representative combinatorial optimization benchmarks. The results demonstrate that our method significantly outperforms state-of-the-art static baselines, exhibiting superior scalability in high-dimensional search spaces. Furthermore, ablation studies confirm that dynamic adaptation fails without grounded perception, often performing worse than static algorithms. This indicates that DyACE's effectiveness stems from the causal alignment between the synthesized logic and the verified gradients of the optimization landscape.
\end{abstract}

\label{sec:intro}

\section{Introduction}

Heuristics serve as a cornerstone approach for navigating the rugged landscapes of complex optimization problems \cite{pearl1983heuristics}. Unlike constructive methods that build solutions sequentially, this study focuses on \textit{perturbative heuristics}---such as Evolutionary Algorithms~\cite{back1997handbook,eiben2015introduction} and Local Search~\cite{lourencco2003iterated}.These methods explore the solution space through iterative transitions, continuously refining solutions towards better regions. However, governed by the ``No Free Lunch'' theorem~\cite{wolpert2002no}, no single universal operator guarantees optimality across all problem topologies. This reality ties a solver's effectiveness to the instance-specific design of its operators. Yet, achieving this tailoring often necessitates manual intervention by domain experts. This requirement creates a design bottleneck, effectively limiting the scalability of general-purpose solvers~\cite{blum2003metaheuristics}.

To address these limitations, AHD emerged, aiming to automate the discovery of algorithms via computational means \cite{burke2013hyper,stutzle2018automated}. Early AHD approaches mainly used Genetic Programming (GP) as hyper-heuristics to evolve heuristic rules for specific instances \cite{koza1994genetic,langdon2013foundations,zhang2023survey}. Although GP advanced automation, its search potential is often restricted by predefined primitive sets and a superficial understanding of code semantics.
This limitation effectively prevents GP from representing complex algorithmic logic \cite{o2010open}.

Recently, the explosive development of Large Language Models (LLMs) \cite{naveed2025comprehensive} has precipitated a paradigm shift, leading to the emerging field of LLM-Driven AHD. Seminal studies have shown that LLMs possess potent program synthesis and logical reasoning capabilities \cite{chen2021evaluating,austin2021program,ma2023eureka,hemberg2024evolving,nasir2024llmatic}, enabling them to function as code-centric optimizers. 

However, a closer look at current LLM-Driven AHD methods reveals a common limitation: they rely on a Static Algorithm Paradigm. most existing methods follow a linear ``offline search-online application'' workflow. They use a large computational budget to synthesize a single optimal algorithm, which is then frozen and deployed \cite{2024FunSearch,2024EoH,ye2024reevo}. This paradigm assumes that a fixed algorithmic structure is sufficient to navigate the entire optimization process. In reality, the perturbative search process is dynamic and non-stationary. 
It typically requires high exploration in early stages to cover the solution space, followed by exploitation for fine-grained convergence \cite{eiben2015introduction,dacosta2008adaptive}. 
Static algorithms ignore these phase transitions, creating a mismatch between their fixed logic and the changing needs of the population. This mismatch often limits scalability in complex landscapes.

To address this, we propose \textit{Dynamic Algorithm Co-evolution (DyACE)}. Unlike conventional methods that view AHD as a static task, we redefine heuristic design as a Non-stationary Bi-level Control Problem using a Receding Horizon Control \cite{dq2009model} architecture. DyACE establishes an online closed-loop system of ``Look-Ahead Rollout Search--Decoupled Meta-Reasoning--Adaptive Actuation.'' By continuously extracting the Search Trajectory Features, DyACE synthesizes time-variant algorithms that dynamically align with the instantaneous geometry of the fitness landscape. This mechanism replaces expensive full-cycle evaluations with efficient, state-aware local look-aheads. It ensures that the algorithmic logic evolves with the solution population, maintaining search effectiveness throughout the entire trajectory. A comprehensive review of related work is provided in Appendix \ref{sec:related_work} due to space constraints.

Specifically, this study makes three primary contributions:

\begin{itemize}
    \item We introduce DyACE, which redefines AHD as a Non-stationary Bi-level Control Problem rather than a static optimization task. By incorporating a Receding Horizon Control architecture, DyACE enables the continuous synthesis of time-variant algorithms that actively adapt to the shifting exploration-exploitation needs of the iterative search trajectory.
    
    \item We establish a closed-loop inference system that transforms the LLM into a grounded meta-controller via \textit{Look-Ahead Rollout Search}. By extracting Search Trajectory Features, the system performs decoupled diagnosis and prescribes phase-specific interventions through three reasoning modes, ensuring alignment with real-time search dynamics.
    
    \item We evaluate DyACE on three representative combinatorial optimization benchmarks. Results show significant scalability and better performance compared to state-of-the-art static methods. Also, ablation studies confirm that dynamic adaptation fails without grounded perception, often performing worse than static algorithms.
\end{itemize}

\section{Problem Reformulation: From Static Optimization to Dynamic Control}
\label{sec:problem_formulation}

\subsection{The Lower Level: Perturbative Search as a Dynamic System}
\label{subsec:lower_level}

We consider a generic optimization problem with a solution space $\mathcal{X}$ and a cost function $f: \mathcal{X} \to \mathbb{R}$. Unlike constructive methods that build a solution sequentially, perturbative heuristics iteratively transform a population of candidate solutions to navigate the fitness landscape.

Formally, we model this search process as a discrete-time dynamic system . Let $\mathcal{P}_t = \{x_1^{(t)}, \dots, x_N^{(t)}\} \in \mathcal{X}^N$ denote the population state at generation $t$. The evolution of this state is governed by a heuristic operator $h$ select from an algorithm space $\Omega$. The state transition dynamics are defined as:
\begin{equation}
    \mathcal{P}_{t+1} = \Phi(\mathcal{P}_t, h_t; \xi_t), \quad t = 0, \dots, T-1
    \label{eq:dynamics}
\end{equation}
Here, $\Phi$ represents the evolutionary update mechanism (including selection and variation), $h_t \in \Omega$ is the heuristic logic applied at step $t$, and $\xi_t$ represents stochastic factors.

The objective is to minimize the cost of the best solution found throughout the trajectory $\mathcal{T} = (\mathcal{P}_0, \dots, \mathcal{P}_T)$. We define the trajectory performance metric as $J(\mathcal{T}) = \min_{t, i} f(x_i^{(t)})$. Eq. (\ref{eq:dynamics}) shows that the final outcome $J(\mathcal{T})$ depends on the sequence of interactions between the operator $h_t$ and the evolving population topology $\mathcal{P}_t$.

\subsection{The Static Upper Level: Time-Invariant Algorithm Optimization}
\label{subsec:static_paradigm}

Existing LLM-based AHD frameworks \cite{2024FunSearch,ye2024reevo} usually formulate the design task as a static bi-level optimization problem. At the upper level, the LLM searches a discrete code space $\mathcal{C}$ to find the best algorithm $c^*$. Let $\Psi: \mathcal{C} \to \Omega$ be the function that maps this code to an executable operator. The static approach aims to solve:
\begin{equation}
    c^* = \operatorname*{argmin}_{c \in \mathcal{C}} \mathbb{E}_{I \sim \mathcal{D}} \left[ J\left( \text{Rollout}(\mathcal{P}_0^{(I)}, \Psi(c)) \right) \right]
    \label{eq:static_obj}
\end{equation}
subject to the time-invariance constraint:
\begin{equation}
    h_t \equiv \Psi(c), \quad \forall t = 0, \dots, T-1
    \label{eq:static_constraint}
\end{equation}
This formulation has a clear limitation. By enforcing Eq. (\ref{eq:static_constraint}), static methods assume that a single ``globally optimal'' operator works well across all search phases—from high-entropy exploration to low-entropy convergence. This assumption ignores the changing needs of the search process, contradicting the No Free Lunch theorem when applied to the temporal dimension.

\label{sec:framework}
\begin{figure*}[t]
    \centering
    \includegraphics[width=\linewidth]{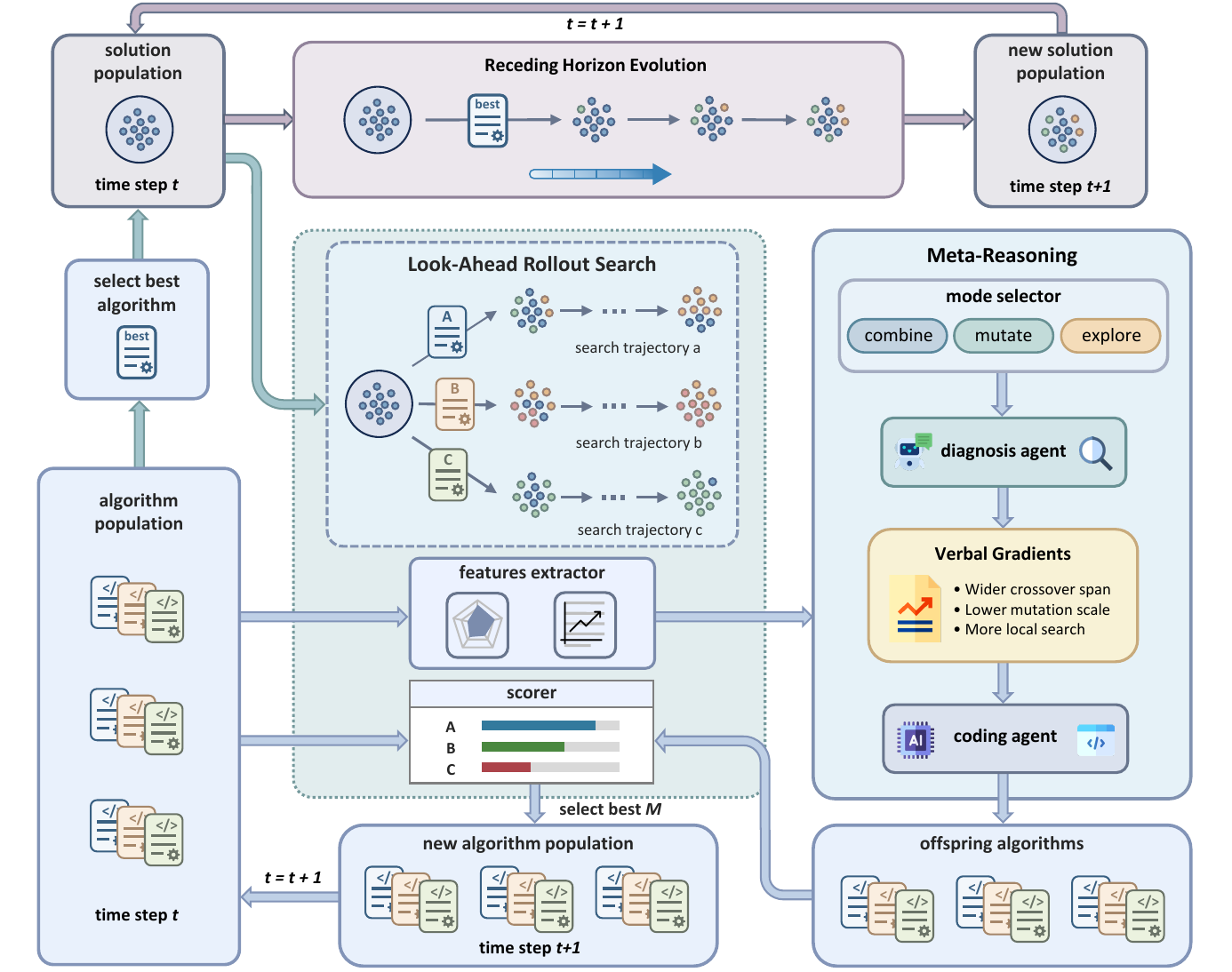}
    \caption{Overview of the DyACE framework. It illustrates the closed-loop interaction between look-ahead perception, meta-reasoning, and adaptive evolution.}
    \label{fig:framework}
\end{figure*}

\subsection{The Dynamic Upper Level: Non-stationary Process Control}
\label{subsec:nbcp}

To address this, we redefined AHD as a Non-stationary Bi-level Control Problem. Instead of freezing the algorithmic logic, we aim to synthesize a \textit{dynamic meta-control policy} $\pi$ that adapts the heuristic operator to the real-time evolutionary state.

Let $\psi: \mathcal{X}^N \to \mathcal{S}$ be an observation function that extracts Search Trajectory Features from the raw population. The state at time $t$ is given by $s_t = \psi(\mathcal{P}_t)$. We relax the constraint in Eq. (\ref{eq:static_constraint}) and define the upper-level problem as learning a policy $\pi: \mathcal{S} \to \Omega$, such that:
\begin{equation}
    h_t = \pi(s_t), \quad t = 0, \dots, T-1
    \label{eq:dynamic_actuation}
\end{equation}
This changes the optimization goal from finding a static code to controlling the optimization trajectory. For a given problem instance, we seek the optimal policy $\pi^*$ that minimizes the expected trajectory cost:
\begin{equation}
    \pi^* = \operatorname*{argmin}_{\pi} \mathbb{E}_{\xi} \left[ J(\mathcal{P}_{0:T}) \mid \mathcal{P}_{t+1} = \Phi(\mathcal{P}_t, \pi(\psi(\mathcal{P}_t)); \xi_t) \right]
    \label{eq:control_obj}
\end{equation}
This formulation transforms heuristic design from an open-loop task into a closed-loop feedback control task. By enabling $h_t$ to vary over time, the meta-controller can switch search operators in response to the changing population structure, theoretically achieving better convergence than static methods.

\section{The DyACE Framework}
\label{sec:Framework}

\subsection{Framework Overview}
\label{subsec:overview}

To address the Non-stationary Bi-level Control Problem formulated in Eq. (\ref{eq:control_obj}), we propose the DyACE framework. Unlike the conventional Static Algorithm Paradigm that uses a fixed, unchanging code $c^*$, DyACE employs a Receding Horizon Control architecture. This approach transforms algorithm design from an open-loop synthesis task into a continuous closed-loop feedback process. By creating an adaptive algorithm trajectory $\tau_S$ (a dynamic sequence of operators $\{h_t\}$), DyACE ensures that the algorithmic logic evolves along with the real-time state topology of the solution population $\mathcal{P}_t$. This effectively aligns the meta-controller's actions with the changing dynamics of the search landscape.

As shown in Figure \ref{fig:framework}, the system operates through three main modules that correspond to the control loop components:
\begin{itemize}
    \item \textbf{Look-Ahead Rollout Search:} This module acts as the perception mechanism. It executes short-term, look-ahead rollouts from the current population $\mathcal{P}_t$ to extract Search Trajectory Features, constructing the state representation $\mathcal{S}_t$ required for Meta-Reasoning.
    
    \item \textbf{Decoupled Meta-Reasoning:} Functioning as the Meta-Controller, the LLM interprets the state $\mathcal{S}_t$ to synthesize a context-specific heuristic operator $h_t = \pi(\mathcal{S}_t)$. This process involves diagnosing the current search status (e.g., stagnation vs. convergence) and prescribing a precise algorithmic intervention.
    
    \item \textbf{Receding Horizon Evolution:} The synthesized operator $h_t$ is applied to the population for a limited horizon $H$, driving the state transition $\mathcal{P}_{t} \to \mathcal{P}_{t+H}$. This receding horizon mechanism ensures that the search strategy is periodically re-calibrated, maintaining alignment with the evolving optimization landscape.
\end{itemize}

\subsection{Look-Ahead Rollout Search}
\label{subsec:probing}

To empower the Meta-Controller with a prognostic understanding of the search dynamics, DyACE employs a Look-Ahead Rollout Search module. This module acts as a sensor, executing short-term rollouts to extract causal signals from the evolutionary trajectory. Specifically, at each decision step $t$, we create parallel shadow environments initialized with the current population $\mathcal{P}_t$. A set of operators are then applied for a limited horizon $T_{probe}$ to generate search trajectories. The information derived from these trajectories is split into two parallel streams to support decoupled reasoning: executing Landscape Feature Extraction to synthesize Search Trajectory Features for state diagnosis, and conducting Iso-State Rollout Evaluation for algorithm scoring.

\subsubsection{Landscape Feature Extraction}

To diagnose the optimization status, this process summarizes the raw search
trajectories into a compact vector of Search Trajectory Features. These features capture the search dynamics through two aspects:

\paragraph{Landscape Kinematics.} We analyze the temporal evolution of the anchor population to infer local landscape geometry. By computing the \textit{velocity} and \textit{acceleration} of the fitness trajectory, we quantify optimization momentum to distinguish between convergence and stagnation. We also monitor the diversity loss rate to see if the search is currently exploring or exploiting .

\paragraph{Operator Telemetry.} Complementing environmental signals, we attribute causality by measuring operator behavior within the rollouts. We calculate \textit{Operator Precision} (the ratio of offspring strictly outperforming parents) to measure alignment with the search direction, and \textit{Operator Impact} (the average fitness gain of successful updates) to differentiate between small local improvements and large structural changes.

\subsubsection{Iso-State Rollout Evaluation}
Alongside feature extraction, we conduct a quantitative assessment to score the algorithms. To reduce noise, we run $M$ independent Monte Carlo rollouts for each algorithm $h_k$ starting from the same $\mathcal{P}_t$. The performance is measured by the average best gap. Let $g(\mathcal{P})$ denote the optimality gap of the best solution in population $\mathcal{P}$ relative to the BKS. The score $J(h_k)$ is calculated as the expected gap at the end of the rollout:
\begin{equation}
    J(h_k) = \frac{1}{M} \sum_{m=1}^{M} g\left(\mathcal{P}_{t+\tau}^{(m)} \big| h_k\right)
    \label{eq:avg_gap}
\end{equation}
A lower $J(h_k)$ means better performance. This metric provides concrete evidence of which algorithm reduces the gap most effectively in the current state.

\subsection{Decoupled Meta-Reasoning and Three Reasoning Modes}
\label{sec:reasoning}

The Inference Engine use the constructed feature $\mathbf{e}_t$ to generate the next algorithm $S_{t+1}$. 
However, directly mapping complex landscape dynamics to code is difficult and can lead to errors or hallucinations.
To solve this, DyACE employs a decoupled cognitive architecture that separates high-level algorithmic planning from low-level implementation.

\subsubsection{Two-Stage Diagnosis-Prescription Architecture}
We break the meta-reasoning process into two sequential steps to ensure a ``Think-then-Code'' approach:

\paragraph{Phase I: The Diagnosis Agent.} This agent act as an analyst. It processes the Search Trajectory Features $\mathbf{e}_t$ but is restricted from writing code. Its main job is to translate landscape dynamics into natural language ``Verbal Gradients'' \cite{ye2024reevo}(e.g., \textit{``Single-block crossover limits mixing; refactor to Two-Point Crossover to exchange disjoint segments and boost structural diversity.''}. This step ensures the system understands why the search is stuck before deciding how to fix it.

\paragraph{Phase II: The Coding Agent.} based on the Verbal Gradients, this agent synthesizes a new offspring algorithm. 
We define the algorithm as a triple $S = \langle \mathcal{D}, \mathcal{C}, \Theta \rangle$, comprising the semantic description $\mathcal{D}$, the executable code $\mathcal{C}$, and the evolutionary hyperparameters $\Theta$.
This structure allows the agent to adjust both logic and parameters. For example, by increasing a diversity weight $\lambda$ along with the code change, the LLM can flatten local optima to help escape traps, ensuring the search parameters align with the intended strategy.

\subsubsection{Three Reasoning Modes}

To implement the reasoning and generation of the new algorithm, the Meta-Reasoning uses one of three reasoning modes:

\paragraph{Combine:} 
This mode uses a pair of parents selected by Structure-Aware Sampling: the primary is chosen by fitness, and the secondary by AST Tree-Edit Distance relative to the first. 
Beyond simple code splicing, the LLM merges the logic of both parents to create a coherent hybrid algorithm .

\paragraph{Mutate:} 
This mode focuses on local optimization within the algorithm space. The LLM refactors code or tunes parameters of a single parent algorithm, aiming to polish its implementation or adjust its search intensity without altering its fundamental algorithmic structure.

\paragraph{Explore:} 
Designed to add structural novelty, this mode use high-temperature sampling to induce Maximal Logical Divergence. 
Rather than refining existing traits, it prompts the LLM to create algorithmic structures that differ significantly from the parents, effectively expanding the search scope.

\begin{table*}[t]
    \centering
    \resizebox{\textwidth}{!}{
    \begin{tabular}{l r rrrrrrrrrrr}
        \toprule
        \textbf{Cases} & 
        \multicolumn{1}{r}{\textbf{Size}} &
        \multicolumn{1}{r}{\textbf{GP}} & 
        \multicolumn{1}{r}{\textbf{GEP}} & 
        \multicolumn{1}{r}{\textbf{DRL\_Chen}} & 
        \multicolumn{1}{r}{\textbf{DRL\_Zhang}} & 
        \multicolumn{1}{r}{\textbf{DRL\_Liu}} & 
        \multicolumn{1}{r}{\textbf{LSO}} & 
        \multicolumn{1}{r}{\textbf{SPT/TWKR}} & 
        \multicolumn{1}{r}{\textbf{FunSearch}} & 
        \multicolumn{1}{r}{\textbf{EoH}} & 
        \multicolumn{1}{r}{\textbf{ReEvo}} & 
        \multicolumn{1}{r}{\textbf{DyACE}} \\
        \midrule
        TA01 & $15\times15$ & 25.67 & 25.67 & 38.99 & 16.41 & 21.20 & 58.98 & 35.17 & 13.81 & 14.13 & 11.78 & \textbf{6.66} \\
        TA02 & $15\times15$ & 25.80 & 19.45 & 31.75 & 24.12 & 14.55 & 41.40 & 23.63 & 13.26 & 11.98 & 10.85 & \textbf{8.36} \\
        TA11 & $20\times15$ & 28.89 & 34.05 & 35.08 & 32.20 & 29.11 & 63.30 & 38.98 & 21.23 & 20.27 & 20.78 & \textbf{18.64} \\
        TA12 & $20\times15$ & 30.87 & 26.70 & 29.11 & 32.04 & 23.77 & 59.99 & 44.04 & 18.95 & 19.17 & 16.75 & \textbf{14.56} \\
        TA21 & $20\times20$ & 27.28 & 27.22 & 30.63 & 37.15 & 27.71 & 61.21 & 34.35 & 20.95 & 21.07 & 19.98 & \textbf{15.23} \\
        TA22 & $20\times20$ & 28.69 & 23.81 & 25.94 & 31.37 & 20.25 & 57.63 & 31.94 & 22.19 & 21.75 & 20.31 & \textbf{16.31} \\
        TA31 & $30\times15$ & 30.78 & 29.20 & 35.03 & 45.41 & 29.08 & 40.48 & 38.04 & 23.24 & 19.73 & 21.71 & \textbf{18.31} \\
        TA32 & $30\times15$ & 34.53 & 34.30 & 37.78 & 33.86 & 23.49 & 47.65 & 40.81 & 27.64 & 25.95 & 25.34 & \textbf{21.92} \\
        TA41 & $30\times20$ & 42.19 & 36.11 & 26.73 & 33.02 & 34.56 & 43.29 & 44.54 & 30.03 & 29.13 & 26.83 & \textbf{21.55} \\
        TA42 & $30\times20$ & 34.38 & 34.90 & 42.59 & 37.53 & 35.42 & 59.83 & 45.22 & 31.34 & 30.82 & 26.90 & \textbf{21.68} \\
        TA51 & $50\times15$ & 30.54 & 32.90 & 36.30 & 30.40 & 30.72 & 39.28 & 36.52 & 25.91 & 21.23 & 26.85 & \textbf{16.96} \\
        TA52 & $50\times15$ & 21.41 & 20.61 & 27.39 & 21.23 & 27.87 & 34.80 & 30.19 & 25.44 & 19.67 & 26.67 & \textbf{14.73} \\
        TA61 & $50\times20$ & 28.49 & 26.99 & 26.67 & 27.41 & 23.71 & 46.03 & 30.82 & 33.23 & 27.86 & 24.02 & \textbf{17.78} \\
        TA62 & $50\times20$ & 26.73 & 29.77 & 29.38 & 29.73 & 23.98 & 46.99 & 36.81 & 37.50 & 23.91 & 25.86 & \textbf{19.17} \\
        TA71 & $100\times20$ & 15.39 & 14.90 & 15.68 & 18.08 & 15.10 & 23.61 & 22.71 & 26.39 & 17.40 & 19.42 & \textbf{11.13} \\
        TA72 & $100\times20$ & 11.48 & 8.57 & 27.81 & 9.92 & 15.85 & 28.82 & 22.58 & 26.98 & 18.34 & 17.47 & \textbf{8.43} \\
        \midrule
        Avg & -- & 28.32 & 26.54 & 31.67 & 28.16 & 24.32 & 47.01 & 33.45 & 24.25 & 20.21 & 19.38 & \textbf{14.73} \\
        \bottomrule
    \end{tabular}
    }
    \caption{Performance Comparison on Taillard JSSP Benchmark. The values denote the optimality gap metric (lower is better). DyACE achieves the best performance across all instances, showing significant advantages in large-scale problems.}
    \label{tab:jssp}
\end{table*}

\section{Experiments}
\subsection{Experimental Settings}

\subsubsection{Benchmarks and Datasets}

We evaluate DyACE on three standard NP-hard Combinatorial Optimization Problems (COPs) to test its performance across different problem types.

\paragraph{Job Shop Scheduling Problem (JSSP).} 
JSSP involves scheduling jobs on machines with precedence constraints to minimize the total completion time (Makespan) .This problem represents a highly constrained discrete optimization challenge.
We use the standard Taillard Benchmark \cite{taillard1993benchmarks}, selecting 16 instances ranging from medium ($15 \times 15$) to large scale ($100 \times 20$). These instances vary in complexity, making them a standard test for measuring the scalability of scheduling algorithms.

\paragraph{Traveling Salesman Problem (TSP).} 
TSP aims to find the shortest Hamiltonian cycle visiting a set of cities exactly once. It assesses the algorithm's capability to navigate continuous Euclidean spaces.
We select 6 classic instances from TSPLIB \cite{reinelt1991tsplib} with scales ranging from 51 to 200 nodes. These instances feature distinct node distributions, testing the robustness of the evolved algorithms in continuous Euclidean spaces.

\paragraph{Capacitated Vehicle Routing Problem (CVRP).} 
CVRP extends TSP by adding vehicle capacity constraints, aiming to minimize travel distance while meeting customer demands .
We employ the CMT dataset from CVRPLIB \cite{uchoa2017newcvrplib}, selecting 6 instances with problem sizes between 50 and 199 nodes. This benchmark tests the algorithm's ability to balance objective optimization with strict feasibility constraints.

\subsubsection{Compared Methods}

To evaluate the efficacy of DyACE, we benchmark it against a diverse set of baselines, including GP, Gene Expression Programming (GEP)~\cite{nie2013reactive}, heuristic dispatching rules (LSO, SPT/TWKR), and representative Deep Reinforcement Learning (DRL) models~\cite{chen2022deep,liu2024dynamic,zhang2020learning}.
Most importantly, we compare against state-of-the-art LLM-based AHD frameworks: FunSearch~\cite{2024FunSearch}, EoH~\cite{2024EoH}, and ReEvo~\cite{ye2024reevo}.
These methods represent the current standard for algorithmic discovery and serve as the primary reference for validating our dynamic co-evolutionary approach .

\subsubsection{Experimental Protocols.}
To ensure a fair comparison, we enforce a unified computational budget of $B=300$ algorithm evaluations and fix the solution population size at $N=100$ for all methods. All LLM-based approaches use \texttt{GPT-4o-mini} as the reasoning engine. 
While static baselines use standard offline evolution, DyACE operates under an online co-evolution protocol.
Comprehensive implementation details, including hyperparameters and the algorithm search space, are provided in Appendix \ref{sec:implementation}.

\begin{figure}[t]
    \centering
    \begin{subfigure}[b]{0.497\linewidth}
        \centering
        \includegraphics[width=\linewidth]{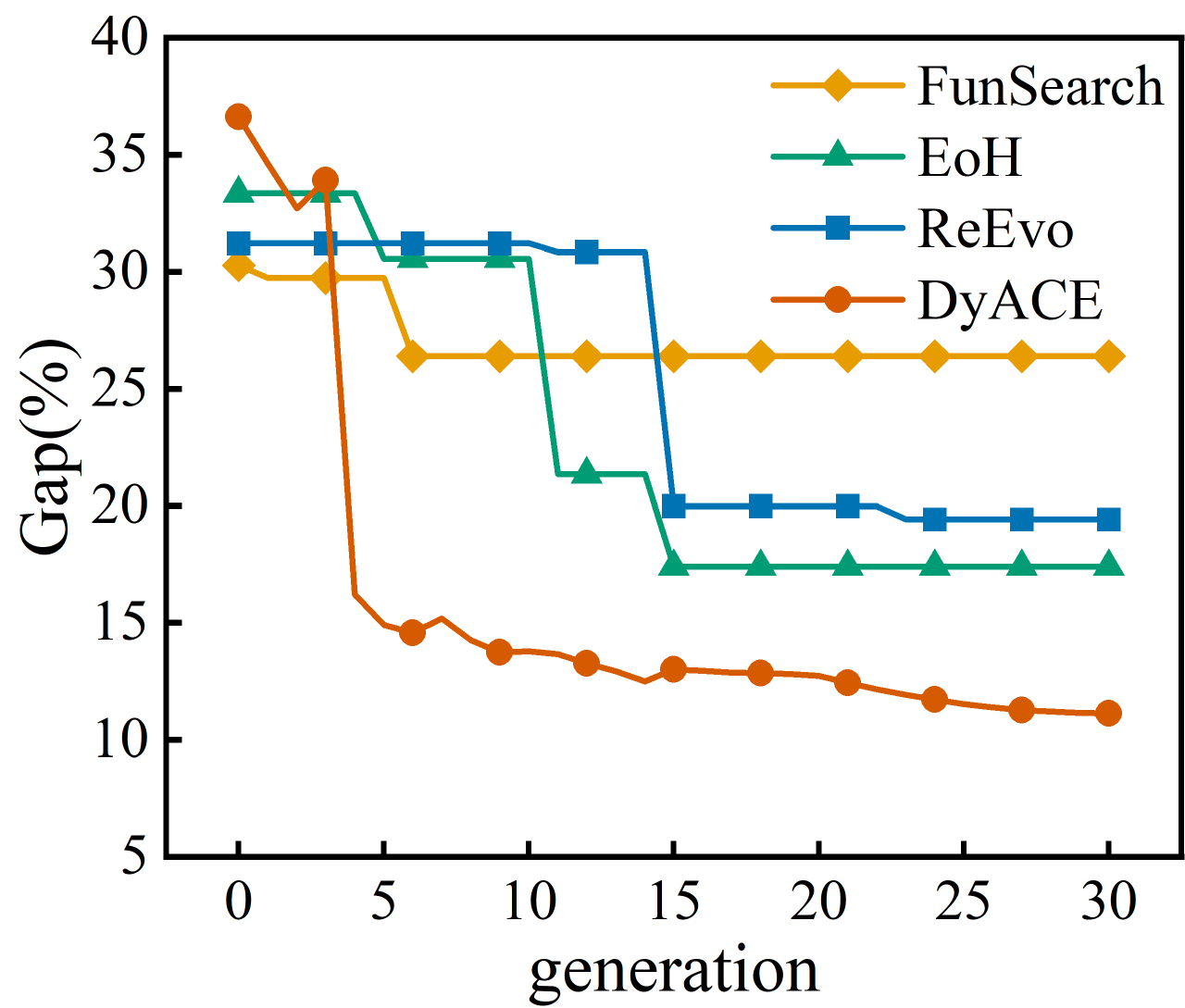}
        \caption{}
        \label{fig:algorithm_evolution}
    \end{subfigure}%
    \hfill 
    \begin{subfigure}[b]{0.497\linewidth}
        \centering
        \includegraphics[width=\linewidth]{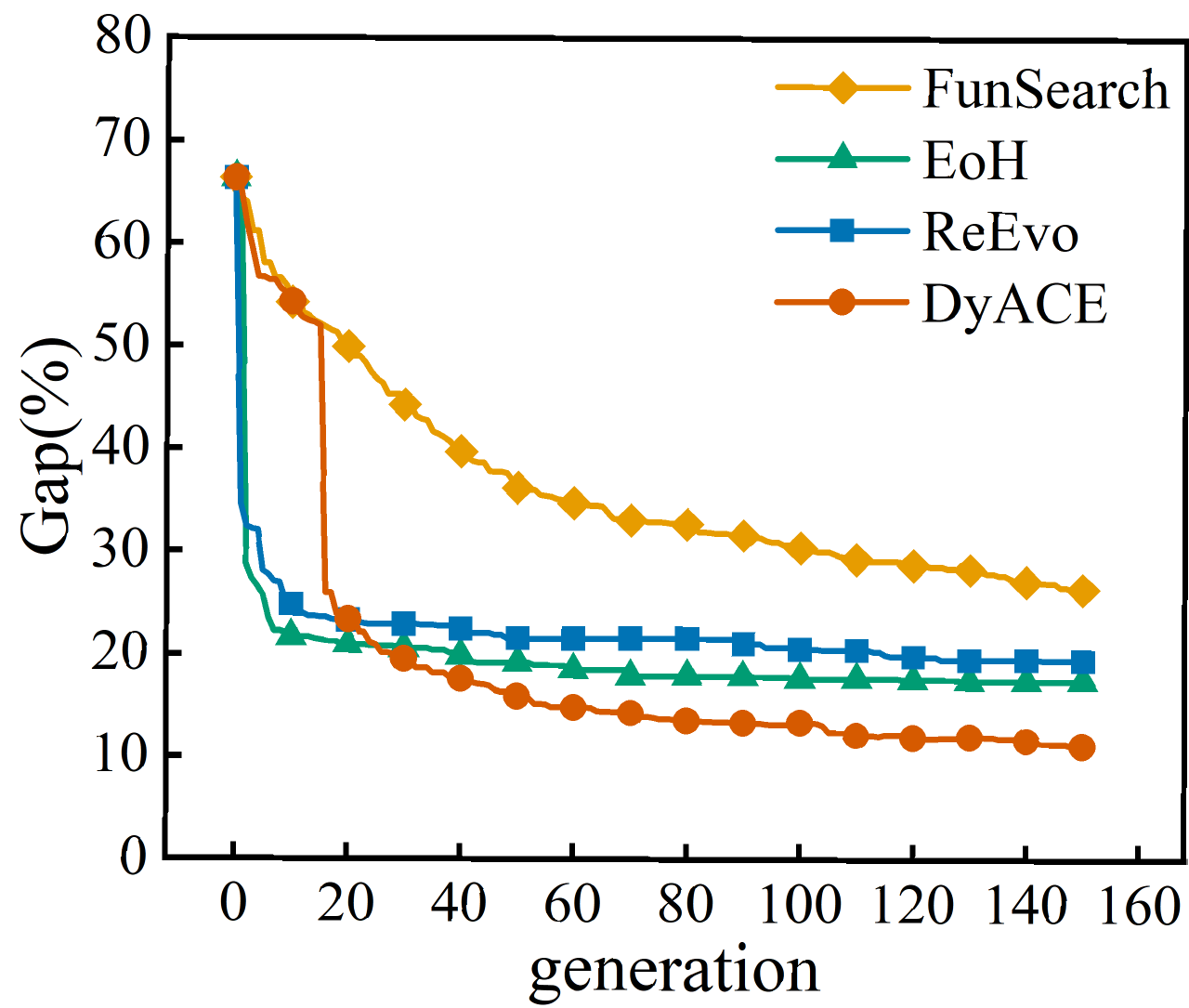}
        \caption{}
        \label{fig:jssp_convergence}
    \end{subfigure}
    
    \caption{Evolutionary iterative process on JSSP instance TA71. (a) The optimality gap of the best algorithm's Monte Carlo evalution  at each generation of the algorithm population. (b) The optimality gap of the best solution throughout the entire search process.}
    \label{fig:dynamics}
\end{figure}

\subsection{Main Results on Job Shop Scheduling (JSSP)}
\label{sec:jssp_results}

Table \ref{tab:jssp} summarizes the results on the Taillard benchmark and demonstrates the effectiveness of the DyACE framework.
On the entire test set, DyACE achieved the lowest optimality gap for every instance, with an average gap of 14.73\%.
This performance is clearly better than the strongest static LLM-based baseline, ReEvo (19.38\%).
Such consistent results suggest that the dynamic co-evolution approach offers a strong advantage over the traditional ``generate-once-and-apply'' static heuristic design, especially in strictly constrained environments.

The difference in performance becomes more obvious as the problems get more complex.
On smaller instances like TA01 ($15\times15$), DyACE reduces the gap to 6.66\%, showing better precision in guiding local search.
the benefit of the Receding Horizon Control architecture is most clear on large-scale instances like TA71 ($100\times20$). 
Here, static methods hit a ``scalability wall,'' where their performance plateaus with gaps between 17.40\% (EoH) and 26.39\% (FunSearch). 
This happens because their fixed operator logic cannot adapt to the deeper and rougher search landscape. 
In contrast, DyACE effectively reduced the gap to 11.13\%. 
This trend confirms that while static heuristics made offline might work for simple problems, the continuous closed-loop feedback of DyACE is necessary to navigate the high-dimensional search spaces of complex scheduling tasks.

To understand the reason for this advantage, we analyzed the evolutionary process on the representative instance TA71.
Figure \ref{fig:dynamics} (a) shows a clear difference in control behavior. 
Static baselines like FunSearch and ReEvo quickly converge to a local optimum in the algorithm space (around generations 5--15) and stop improving at the meta-level. 
This stagnation reflects a key limit of the static paradigm: because they evolve algorithms based only on the final solution quality starting from a poor initial population, these methods cannot see the intermediate needs of the search process. 
As a result, they search the vast program space with only a simple, delayed reward signal, often getting trapped in logic that is not good enough for late-stage exploitation.

In contrast, DyACE keeps reducing the algorithm optimality gap throughout the entire process (Figure \ref{fig:dynamics} (a)). 
This trajectory shows that the Receding Horizon Control loop is working effectively: as the solution population evolves, the Look-Ahead Rollout sensor detects changes in the landscape . This prompts the Meta-Controller to create new, phase-specific operators.
This continuous adaptation leads directly to better solutions, as shown in Figure \ref{fig:dynamics} (b). 
While baselines show flat convergence curves, which means they are stuck, DyACE shows a staircase convergence pattern. 
The inner loop (solution evolution) improves the population under the current algorithm, and the outer loop (algorithm update) periodically renews the search momentum with updated logic. 
This combination allows the population to break through local optima and achieve much deeper convergence.

\subsection{Generalization to Routing Problems}
\label{subsec:routing-results}

To assess the generalizability of DyACE, we extended our evaluation to routing domains: the TSP and the CVRP. 
Specifically, to isolate the algorithmic discovery capability from domain-specific engineering, we restricted the search space to pure evolutionary frameworks (using only Crossover and Mutation), excluding auxiliary local search operators. This setting creates a challenging benchmark that minimizes the influence of underlying solver heuristics, thereby highlighting the quality of the evolved logic.

\subsubsection{Traveling Salesman Problem (TSP)}
\label{subsubsec:tsp}

Table \ref{tab:tsp_original} presents the optimality gap (\%) on standard TSPLIB instances. Under the strict ``Crossover+Mutation only'' constraint, the results reveal a clear divergence in scalability between the dynamic framework and static baselines.

\begin{table}[t]
    \centering

    \resizebox{\columnwidth}{!}{
    \begin{tabular}{lrrrrrr}
        \toprule
        \textbf{Method} & \textbf{eli51} & \textbf{st70} & \textbf{rd100} & \textbf{bier127} & \textbf{u159} & \textbf{kroB200} \\
        \midrule
        FunSearch & 2.347 & 9.926 & 50.834 & 61.979 & 157.129 & 218.215 \\
        EoH       & 0.469 & 3.704 & 13.654 & 5.249  & 3.092   & 19.900 \\
        ReEvo     & 0.00  & 0.00  & 0.607  & 1.713  & 2.583   & 16.194 \\
        \textbf{DyACE (Ours)} & \textbf{0.00}  & \textbf{0.00}  & \textbf{0.00}   & \textbf{0.00}   & \textbf{0.00}    & \textbf{6.376}  \\
        \bottomrule
    \end{tabular}
    }
    \caption{Optimality Gap (\%) on TSPLIB Instances using only crossover and mutation operators. DyACE maintains high solution quality as problem scale increases.}
    \label{tab:tsp_original}
\end{table}

On small-scale instances (eli51, st70), both ReEvo and DyACE successfully identify the global optimum (0.00\% gap), confirming that current LLM-based methods can synthesize valid permutation logic. However, the absence of local search operators exposes the limitations of static heuristics as problem complexity increases. On medium-scale instances like rd100 and u159, ReEvo fails to maintain optimality, yielding gaps ranging from 0.6\% to 2.5\%. In contrast, DyACE consistently converges to the known optimum (0.00\%), demonstrating that its dynamically evolved operators can navigate the search space effectively even without auxiliary exploitation mechanisms.

This performance gap becomes most distinct on the largest tested instance, kroB200. Here, the high-dimensional search space causes static methods to stagnate: FunSearch and EoH fail to converge (gaps $>19\%$), and ReEvo degrades to a 16.194\% gap. Conversely, DyACE restricts the gap to 6.376\%. This empirical evidence suggests that while static AHD methods falter when facing complex problems, DyACE's dynamic adaptation provides a critical robustness, allowing the solver to adjust its exploration strategy in real-time to escape the topological traps typical of large Euclidean spaces.

\subsubsection{Capacitated Vehicle Routing Problem (CVRP)}
\label{subsubsec:cvrp}

The CVRP introduces capacity constraints, requiring the solver to simultaneously optimize route length and satisfy demand limitations. Table \ref{tab:cvrp_original} summarizes the performance comparison on the CVRPLIB (CMT set).DyACE achieved the lowest optimality gap across all six evaluated instances, demonstrating a level of consistency that static baselines failed to match.

\begin{table}[b]
    \centering
    \resizebox{\columnwidth}{!}{
    \begin{tabular}{lrrrrrr}
        \toprule
        \textbf{Method} & \textbf{CMT1} & \textbf{CMT2} & \textbf{CMT3} & \textbf{CMT11} & \textbf{CMT4} & \textbf{CMT5} \\
        \midrule
        FunSearch & 25.580 & 47.330  & 74.850  & 128.464 & 134.474 & 166.816 \\
        EoH       & 17.044 & 33.493  & 19.148  & 33.013  & 20.615  & 168.837 \\
        ReEvo     & 12.570 & 41.270  & 17.120  & 21.455  & 18.953  & 12.963  \\
        \textbf{DyACE (Ours)} & \textbf{3.171}  & \textbf{28.441}  & \textbf{16.514}  & \textbf{15.759}  & \textbf{16.138}  & \textbf{12.010}  \\
        \bottomrule
    \end{tabular}
    }
    \caption{Optimality Gap (\%) on CVRPLIB (CMT) Instances. The results show that DyACE consistently leads across all tested cases.}
    \label{tab:cvrp_original}
\end{table}

The performance advantage is most evident on instance CMT1, where DyACE reduced the gap to 3.17\%, significantly outperforming the closest competitor, ReEvo (12.57\%). 
More importantly, DyACE exhibits superior stability. 
Static methods show high performance variance depending on the instance; for example, on CMT2, ReEvo degraded to a 41.27\% gap, falling behind EoH (33.49\%). 
In contrast, DyACE maintained its lead with a gap of 28.44\%. 
Even on instances where the margin narrowed, such as CMT3 and CMT5, DyACE remained the only method to hold the top position across the entire suite. 
This stability confirms that the non-stationary control mechanism allows the algorithm to adapt robustly to the diverse constraint geometries of different problem instances.

\subsection{Ablation Study: Deconstructing the Control Loop}
\label{subsec:ablation}

To explicitly isolate the contributions of the non-stationary adaptation loop and the look-ahead perception mechanism within the DyACE framework, we conducted a rigorous ablation study on three representative JSSP instances of varying scales. We compared the full DyACE framework against three degraded variants: \textit{DyACE-Static}, which removes the Receding Horizon Control loop to freeze the heuristic logic then evolve the solution population; \textit{DyACE-Blind}, which retains the online adaptive loop but excises the feature extractor module, forcing the Meta-Controller to operate on open-loop feedback without the Search Trajectory Feature vector; and \textit{DyACE-w/o-Both}, a minimal baseline where both mechanisms are removed.

\begin{table}[t]
    \centering
    \resizebox{\columnwidth}{!}{
    \begin{tabular}{lrrr}
        \toprule
        \textbf{Method} & \textbf{ta21} ($20\times20$) & \textbf{ta51} ($50\times15$) & \textbf{ta71} ($100\times20$) \\
        \midrule
        \textbf{DyACE-w/o-both}     & 20.95          & 26.41          & 20.20          \\
        \textbf{DyACE-Blind}        & 17.36          & 21.45          & 19.60          \\
        \textbf{DyACE-Static}       & 16.63          & 19.17          & 17.94          \\
        \textbf{DyACE (Full)}       & \textbf{15.23} & \textbf{16.96} & \textbf{11.13} \\
        \bottomrule
    \end{tabular}
    }
    \caption{Ablation study on JSSP instances. The results confirm that both the co-evolution loop and search trajectory features are essential for effective performance.}
    \label{tab:ablation}
\end{table}

Table \ref{tab:ablation} details the impact of each component. The comparison between the full model and DyACE-Static demonstrates the necessity of non-stationary control, especially as problem complexity increases.
On the small-scale instance ta21, the performance difference is modest, with the static baseline trailing by approximately 1.4\%. However, on the large-scale ta71, this gap widens significantly, where DyACE-Static degrades to a gap of 17.94\% while the full model suppresses it to 11.13\%. This trend confirms that while a time-invariant heuristic might work for simple topologies, the rugged, high-dimensional landscapes of large-scale problems require a continuous, phase-dependent shift in search logic.

More importantly, the behavior of the \textit{DyACE-Blind} variant illuminates the fundamental hazard of uninformed adaptation. On the challenging ta71 instance, \textit{DyACE-Blind} ($19.60\%$) not only underperforms the full model but yields results inferior to the static baseline ($17.94\%$). This phenomenon underscores a core control-theoretic principle: uninformed feedback often introduces instability. Without the causal grounding provided by the Trajectory Features, the adaptive adjustments from the LLM effectively become a random walk within the operator space, disrupting the optimization momentum rather than guiding it. Consequently, the Look-Ahead Rollout Search module is proven to be not merely an auxiliary feature but an essential requirement that transforms the meta-reasoning process from an open-loop guess into a principled, evidence-driven control trajectory.

\section{Conclusion and Future Works}
\label{sec:conclusion_and_future_works}

\subsection{Conclusion}
\label{sec:conclusion}

This work challenges the standard ``Static Algorithm Paradigm'' in Automated Heuristic Design, arguing that it is insufficient for the changing dynamics of combinatorial search.
By reframing the problem as a Non-stationary Bi-level Control Problem, we introduced DyACE, a framework that changes algorithm design from a one-time synthesis task into a continuous process of adaptive control.

Our experiments on JSSP, TSP, and CVRP show that large-scale optimization is not a fixed target but a shifting process that requires real-time adjustments. DyACE achieves this by combining Look-Ahead Rollout Search with a Receding Horizon Control architecture. This combination allows the system to use features from the search process—both the landscape geometry and operator performance—to actively update the search strategy. Most importantly, our ablation study provides a key finding: dynamic adaptation without grounded perception is harmful, as shown by the failure of the ``blind'' control variants. Therefore, the success of DyACE comes strictly from the alignment between the LLM's reasoning and the verified state of the solution population.

Finally, this work suggests a shift in research focus: moving from discovering a single optimal algorithm to designing autonomous agents capable of managing the optimization trajectory

\subsection{Limitations and Future Work}
\label{sec:limitations}

While DyACE shows the value of closed-loop algorithmic control, there are several areas for improvement.

\paragraph{Inference Latency.}
The main bottleneck is the time cost of the Meta-Controller, which limits its use in real-time dispatching scenarios. Although the Receding Horizon Control architecture helps by separating decision steps from population updates, the computational cost is still significant. Future work could explore a hierarchical structure, using smaller, faster models for frequent adjustments and reserving large foundation models for major strategic changes.

\paragraph{Cross-Domain Transfer.}
Currently, DyACE starts from scratch for each instance. However, the natural language representation of Verbal Gradients offers a unique medium for Cross-Domain Transfer. Future work will investigate how to store these insights to create a generalist agent. This would allow the system to transfer successful search patterns (like escape mechanisms learned in TSP) to related domains like CVRP, reducing the need to learn everything anew.

\paragraph{Scientific Discovery.}
Beyond performance, the trajectory of evolved algorithms ($\tau_{S}^{*}$) contains valuable information about the problem structure. While the current system uses this data implicitly for control, explicitly analyzing these trajectories offers a path toward Automated Scientific Discovery. Future research could use secondary LLM agents to analyze the sequence of effective operators, distilling them into human-readable design principles or mathematical conjectures. This would transform AHD from a black-box optimization tool to an instrument for theoretically understanding the intrinsic complexity of these problems.

\bibliographystyle{named}
\bibliography{refs}  

\clearpage
\appendix
\onecolumn
\setcounter{page}{1}

\section{Related Work}
\label{sec:related_work}

\subsection{Perturbative Heuristics}
\label{subsec:perturbative_heuristics}

Heuristic methods are generally divided into two categories: constructive and perturbative \cite{burke2013hyper,gendreau2010metaHhandbook}. While constructive heuristics build a solution sequentially from scratch (e.g., dispatching rules in JSSP), Perturbative Heuristics operate on complete, instantiated candidate solutions. This category includes a range of solvers, from classic Iterative Improvement and Stochastic Local Search \cite{hoos2018stochastic} to population-based Evolutionary Algorithms.

The main feature of these methods is their use of search operators —such as the $k$-opt moves in TSP \cite{lin1973effective} or crossover in EAs—to transform the current state into a neighboring state. Unlike constructive processes that are typically static and one-pass, perturbative heuristics engage in a continuous navigation of the fitness landscape. Therefore, the success of these solvers depends not on a fixed decision rule, but on the ability of their operators to adjust the search trajectory, balancing local exploitation with global exploration. This distinction shows that designing for perturbative heuristics is fundamentally a problem of controlling search dynamics, rather than just synthesizing static mapping rules.

\subsection{Automated Heuristic Design (AHD)}
\label{subsec:ahd}

To overcome the limits of manual algorithm design, AHD aims to automate the construction of solvers \cite{stutzle2018automated,burke2013hyper}. Early work focused on Algorithm Configuration (AC), which optimizes parameters within fixed frameworks using tools like ParamILS \cite{hutter2009paramils} and irace \cite{lopez2016irace}. Subsequently, Hyper-heuristics (HH)---particularly GP---advanced to evolving executable heuristics directly. While effective at creating constructive rules for combinatorial domains \cite{stutzle2018automated,pillay2018hyper}, GP remains encumbered by the semantic gap when handling complex perturbative logic \cite{o2010open}. Its random subtree operations often produce code that is syntactically correct but meaningless . This leads to code bloat and poor readability, making it difficult to discover high-level search operators \cite{mei2022explainable}.

\subsection{LLM-Driven Automated Heuristic Design}
\label{subsec:llm_ahd}

Large Language Models have initiated a paradigm shift in AHD, bridging the semantic gap via superior code generation. Following early black-box explorations \cite{yang2023large,fernando2023promptbreeder,liu2024large}, the field moved toward a white-box evolutionary approach. Key frameworks like \textit{FunSearch} \cite{2024FunSearch} and \textit{EoH} \cite{2024EoH} utilize LLMs as stochastic mutation operators to evolve executable heuristics \cite{2024LLaMEA}. Later studies improved this process by incorporating verbal reflection (\cite{ye2024reevo}), co-evolving test cases (\cite{li2025cocoevo}), or even fine-tuning the model parameters (\cite{huang2025calm}) to better align generation with high-performing patterns.

However, these methods still rely on a ``Static Algorithm Paradigm''. Whether using a frozen, reflective, or fine-tuned model, the final output is a fixed, time-invariant algorithm meant for general use. While effective, this static approach often lacks the flexibility to respond to the non-stationary dynamics inherent in perturbative search, where the ideal optimization logic may need to vary significantly between early-stage exploration and late-stage exploitation.

\section{Implementation Details}
\label{sec:implementation}

\subsection{Hardware and General Settings.}
We executed experiments on a single node equipped with an AMD EPYC 9654 CPU and employed \texttt{GPT-4o-mini} as the underlying reasoning engine for all LLM-based methods. For the traditional GP and GEP baselines, we followed the experimental setup described in \cite{huang2024automatic}. To ensure a fair comparison at the solution level, we standardized the lower-level configuration across all methods. Specifically, the solution population size was fixed at $N=100$, and we initialized these populations randomly to guarantee an unbiased starting point. The universal objective for all method was to minimize the optimality gap of the best-found solution .

\subsection{Definition of Algorithm Space.}
The target algorithm structures differ by domain to accommodate specific problem characteristics. For the Job Shop Scheduling Problem, we aimed to evolve a Memetic Algorithm framework comprising three operators: Crossover, Mutation, and Local Search. This design leverages local search to accelerate convergence while using evolutionary operators to escape local optima. Conversely, for the Traveling Salesman Problem and Capacitated Vehicle Routing Problem, we restricted the design space to a pure Evolutionary Algorithm framework containing only Crossover and Mutation operators. By excluding auxiliary local search components, this setting imposes a stricter challenge on the AHD methods and forces them to rely solely on the evolved evolutionary logic to navigate the solution space.

\subsection{Evaluation Protocol and Budget Constraints.}
Unlike constructive heuristics that generate a single solution sequentially, perturbative heuristics rely on the iterative evolution of a population. Evaluating the quality of such heuristics requires running Monte Carlo simulations of the entire evolutionary process, which incurs a heavy computational cost. This characteristic makes large-scale training across massive datasets impractical for the LLM-based Automated Perturbative Heuristic Design task. Consequently, we adopted an instance-specific evolution approach. To balance performance with computational feasibility, we set a strict budget of $B=300$ algorithm evaluations for each problem instance. The specific operational protocols for the baselines and our method are detailed below.

\paragraph{Baselines (Offline Algorithm Search).}
Comparison methods, including FunSearch, EoH, and ReEvo, operate under an offline algorithm search mechanism. These methods first expend the budget of 300 evaluations to evolve an algorithm population. Once this offline phase converges, the system freezes the best-discovered algorithm and applies it to the solution population for the full 150 generations to measure the final optimality gap. In this phase, each algorithm evaluation involves a full rollout (evolving from generation 0 to 150), with a time limit of 5 minutes per evaluation. Regarding specific configurations, we configured FunSearch with a 5-island evolutionary model to accommodate the tight budget, sampling 2 parents for each evolution step. For EoH, we retained the default parameters as specified in the original literature \cite{2024EoH}. For ReEvo, we set the algorithm population size to $M=10$ and evolved it for 30 generations to identify the optimal algorithm within the budget.

\paragraph{DyACE (Online Dynamic Co-evolution).}
In contrast to the offline approach, DyACE embeds the solution evolution directly within the algorithm optimization loop. Specifically, for every generation of algorithm evolution, the solution population advances by 5 generations. Since the solution landscape is non-stationary, the system must re-evaluate parent algorithms at each step to reflect their performance on the current topology. To offset this additional evaluation overhead while maintaining the unified budget, we constrained the algorithm population size to $M=5$. The algorithm layer evolves for 30 generations, which synchronizes exactly with the solution layer's 150-generation horizon ($30 \text{ steps} \times 5 \text{ generations}$). Crucially, the Look-Ahead Rollout Search mechanism allows DyACE to utilize a short-horizon rollout of 30 generations instead of a full 150-generation rollout. This design significantly reduces the computational cost and allows us to lower the time limit per evaluation to 2 minutes.

\section{Prompt Design Details}
\label{app:prompts}

To ensure reproducibility, we provide the core prompt templates used in the Decoupled Meta-Reasoning module. The interaction involves two distinct agents: the Diagnosis Agent and the Coding Agent.

\subsection{Diagnosis Agent Prompt}
The Diagnosis Agent receives the Search Trajectory Features and is tasked with generating Verbal Gradients. It does not generate code but analyzes the search status.


\begin{promptbox}[Combine Prompt for Diagnosis Agent]
You are an expert in \{problem type\} optimization algorithms, tasked with analyzing the advantages of the parent algorithms and propose a fusion optimization method.

\vspace{0.5em}
Algorithm 1 : \{parent1\}

search trajectory on solution population:

\{parent1\_feature\}

\vspace{0.5em}
Algorithm 2 : \{parent2\}

search trajectory on solution population:

\{parent2\_feature\}

\vspace{0.5em}
Conduct a thorough analysis and propose optimization suggestions:

1. Compare their search trajectory performance.

2. Identify each algorithm’s key strengths and weaknesses based on the search trajectory and parameter settings.

3. Propose specific directions for combining the two parent strategies.

\vspace{0.5em}
Must output in the following format:

\texttt{<analysis>} Detailed analysis \texttt{<analysis>}

\texttt{<direction>} Specific Improvement Directions \texttt{</direction>}

\end{promptbox}

\vspace{0cm} 
\captionof{promptcap}{The combine mode prompt used for the Diagnosis Agent.}
\label{prompt:analyze_combine} 
\vspace{1em} 

\begin{promptbox}[Mutate Prompt for Diagnosis Agent]
You are an expert in \{problem type\} optimization algorithms. Please analyze the parent algorithm and provide specific improvement directions.

\vspace{0.5em}
Parent algorithm : \{parent\_algorithm\}

search trajectory on solution population:

\{parent\_feature\}

\vspace{0.5em}
Conduct a thorough analysis and propose optimization suggestions:

1. Analyse parent algorithm's search trajectory performance.

2. Identify the main strengths and weaknesses of the parent algorithm based on the search trajectory and parameter settings.

3. Propose specific local optimization directions that make targeted adjustments to the parent algorithm.

\vspace{0.5em}
Must output in the following format:

\texttt{<analysis>} Detailed analysis \texttt{<analysis>}

\texttt{<direction>} Specific Improvement Directions \texttt{</direction>}

\end{promptbox}

\vspace{0cm} 
\captionof{promptcap}{The mutate mode prompt used for the Diagnosis Agent.}
\label{prompt:analyze_mutate} 
\vspace{1em} 

\begin{promptbox}[Explore Prompt for Diagnosis Agent]
You are an expert in \{problem type\} optimization algorithms, tasked with analyzing a parent algorithm and proposing innovative, distinct optimization approaches.

\vspace{0.5em}
Parent algorithm : \{parent\_algorithm\}

search trajectory on solution population:

\{parent\_feature\}

\vspace{0.5em}
Conduct a thorough analysis and propose optimization suggestions:

1. Analyse parent algorithm's search trajectory performance.

2. Identify the main strengths and weaknesses of the parent algorithm based on the search trajectory and parameter settings.

3. Propose specific exploratory algorithm optimization directions that differ substantially from the parent.

\vspace{0.5em}
Must output in the following format:

\texttt{<analysis>} Detailed analysis \texttt{<analysis>}

\texttt{<direction>} Specific Improvement Directions \texttt{</direction>}

\end{promptbox}

\vspace{0cm} 
\captionof{promptcap}{The explore mode prompt used for the Diagnosis Agent.}
\label{prompt:analyze_explore} 
\vspace{1em} 

\subsection{Coding Agent Prompt}
The Coding Agent use the Verbal Gradients from the Diagnosis Agent to synthesize the executable Python code for the next heuristic operator.


\begin{promptbox}[Initialize Prompt for Coding Agent]
You are a senior Python algorithm engineer tasked with generating directly executable algorithms for the \{problem\_type\}.

\vspace{0.5em}
Seed Algorithm : \{parent\}

search trajectory on solution population:

\{parent\_feature\}

\vspace{0.5em}
Output format: Only three sections, output text directly:

   Description is wrapped in \texttt{<description>}...\texttt{</description>}
   
   Parameters are wrapped in \texttt{<parameter>}...\texttt{</parameter>}
   
   Code is wrapped in \texttt{<code>}...\texttt{</code>}

\vspace{0.5em}
\texttt{<description>}

(First line: concise Algorithm Title)

(Then 2–3 short paragraphs explaining the main ideas.)

\texttt{</description>}

\texttt{<parameter>}

\{parameter\_to\_evolve\}

\texttt{</parameter>}

\texttt{<code>}

\{algorithm\_to\_evolve\}

\texttt{</code>}

\end{promptbox}

\vspace{0cm} 
\captionof{promptcap}{The initialize mode prompt used for the Coding Agent.}
\label{prompt:init} 
\vspace{1em} 


\begin{promptbox}[Combine Prompt for Coding Agent]
You are a senior Python metaheuristic engineer. Task: Combine the advantages of two parent optimization algorithms to produce a new directly executed algorithm code.

\vspace{0.5em}
Algorithm 1 : \{parent1\}

search trajectory on solution population:

\{parent1\_feature\}

\vspace{0.5em}
Algorithm 2 : \{parent2\}

search trajectory on solution population:

\{parent2\_feature\}

\vspace{0.5em}
\{optimization\_direction\}

\vspace{0.5em}
Output format: Only three sections, output text directly

   Description is wrapped in \texttt{<description>}...\texttt{</description>}
   
   Parameters are wrapped in \texttt{<parameter>}...\texttt{</parameter>}
   
   Code is wrapped in \texttt{<code>}...\texttt{</code>}

\vspace{0.5em}
\texttt{<description>}

(First line: concise Algorithm Title)

(Then 2–3 short paragraphs explaining the main ideas.)

\texttt{</description>}

\texttt{<parameter>}

\{parameter\_to\_evolve\}

\texttt{</parameter>}

\texttt{<code>}

\{algorithm\_to\_evolve\}

\texttt{</code>}

\end{promptbox}

\vspace{0cm} 
\captionof{promptcap}{The combine mode prompt used for the Coding Agent.}
\label{prompt:combine} 
\vspace{1em} 


\begin{promptbox}[Mutate Prompt for Coding Agent]
You are a senior Python metaheuristic engineer. Task: Make incremental improvements to the reference algorithm, outputting a directly executable new algorithm.

\vspace{0.5em}
Algorithm to improve : \{parent\}

search trajectory on solution population:

\{parent\_feature\}

\vspace{0.5em}
\{optimization\_direction\}

\vspace{0.5em}
Output format: Only three sections, output text directly

   Description is wrapped in \texttt{<description>}...\texttt{</description>}
   
   Parameters are wrapped in \texttt{<parameter>}...\texttt{</parameter>}
   
   Code is wrapped in \texttt{<code>}...\texttt{</code>}

\vspace{0.5em}
\texttt{<description>}

(First line: concise Algorithm Title)

(Then 2–3 short paragraphs explaining the main ideas.)

\texttt{</description>}

\texttt{<parameter>}

\{parameter\_to\_evolve\}

\texttt{</parameter>}

\texttt{<code>}

\{algorithm\_to\_evolve\}

\texttt{</code>}

\end{promptbox}

\vspace{0cm} 
\captionof{promptcap}{The mutate mode prompt used for the Coding Agent.}
\label{prompt:mutate} 
\vspace{1em} 


\begin{promptbox}[Explore Prompt for Coding Agent]
You are a senior Python metaheuristic engineer. Task: Propose a completely new JSSP optimization algorithm with different ideas from the reference algorithm.

\vspace{0.5em}
Reference Algorithm : \{parent\}

search trajectory on solution population:

\{parent\_feature\}

\vspace{0.5em}
\{optimization\_direction\}

\vspace{0.5em}
Output format: Only three sections, output text directly

   Description is wrapped in \texttt{<description>}...\texttt{</description>}
   
   Parameters are wrapped in \texttt{<parameter>}...\texttt{</parameter>}
   
   Code is wrapped in \texttt{<code>}...\texttt{</code>}

\vspace{0.5em}
\texttt{<description>}

(First line: concise Algorithm Title)

(Then 2–3 short paragraphs explaining the main ideas.)

\texttt{</description>}

\texttt{<parameter>}

\{parameter\_to\_evolve\}

\texttt{</parameter>}

\texttt{<code>}

\{algorithm\_to\_evolve\}

\texttt{</code>}

\end{promptbox}

\vspace{0cm} 
\captionof{promptcap}{The explore mode prompt used for the Coding Agent.}
\label{prompt:explore} 
\vspace{1em} 

\begin{figure*}[t]
    \centering
    \includegraphics[width=\linewidth]{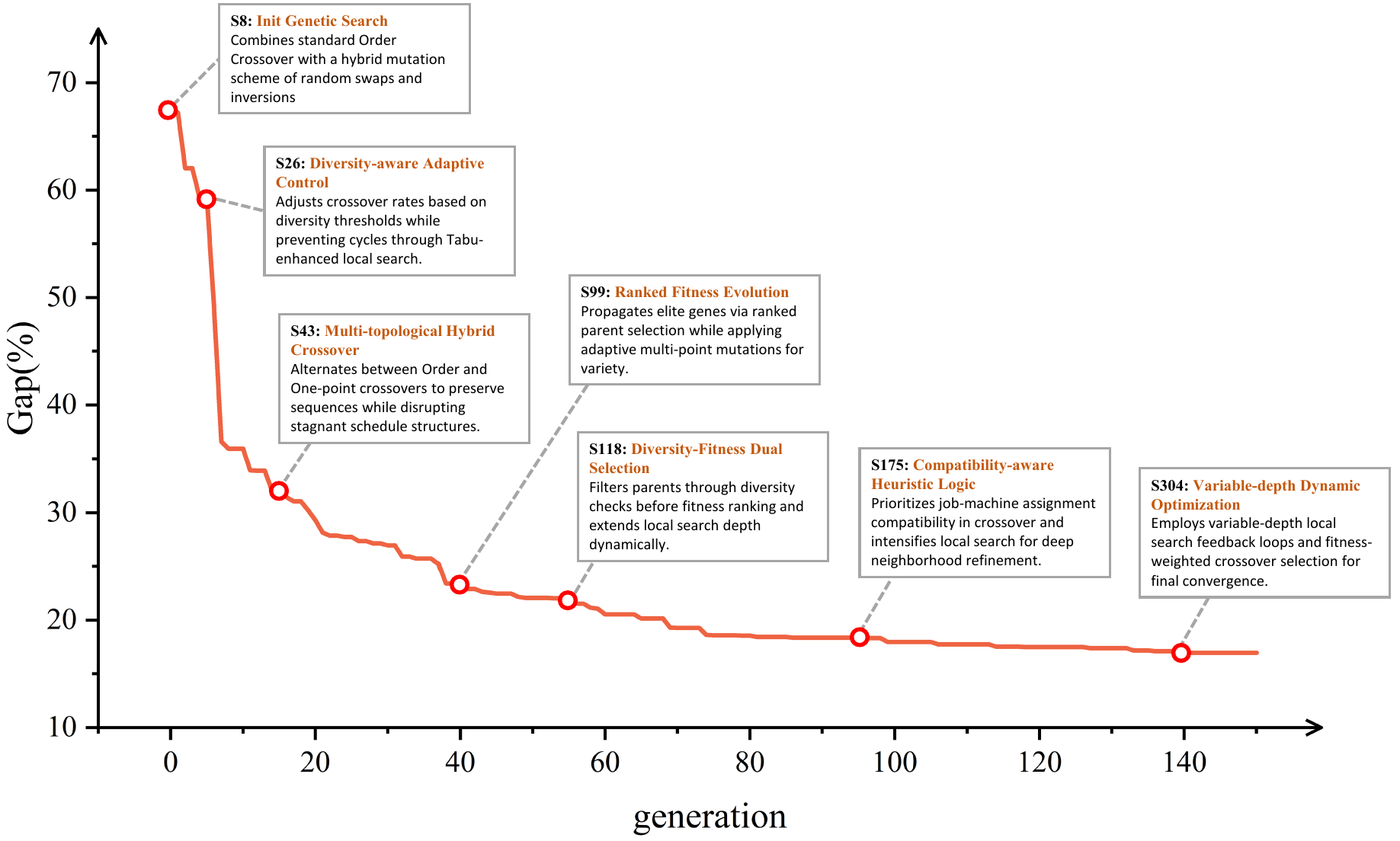}
    \caption{Case study of algorithmic evolution on JSSP instance ta51. The markers indicate key transitions in heuristic logic synthesized to address specific search stages.}
    \label{fig:casestudy}
\end{figure*}

\section{Case Study: Evolutionary Trajectory on JSSP Instance ta51}
\label{app:case_study_detail}
To demonstrate the causal link between landscape dynamics and DyACE's decisions, we analyze the evolutionary trajectory on the JSSP instance ta51 ($50 \times 15$). Figure \ref{fig:casestudy} illustrates the convergence curve, which correlates directly with the specific algorithmic evolutions detailed below.

\subsection{Initial Exploration and Adaptive Divergence}The search process begins with a rapid initial descent led by Algorithm S8 in Generation 0. This algorithm establishes a robust baseline by combining standard Order Crossover (OX) with a balanced mutation scheme that utilizes both random swaps and inversions. However, the system quickly encounters a resistance level where improvements slow down. By Generation 5, the Diagnosis Agent identifies that the current search suffers from low population diversity. In response, DyACE synthesizes Algorithm S26, which introduces a diversity-aware adaptive control mechanism. This algorithm adjusts crossover rates based on a diversity threshold of $0.0917$ and incorporates Tabu Search logic into the local search phase to prevent the population from cycling back to previously evaluated schedules. Additionally, it adopts a more aggressive multi-operation swap mutation that modifies three job positions simultaneously to force exploration into new regions of the solution space.

\subsection{Structural Hybridization and Selection Optimization}As the evolution progresses to Generation 15, the population faces a structural bottleneck where standard operators fail to locate new promising regions. To address this, DyACE transitions to Algorithm S43, which features a Multi-topological Hybrid Crossover. Instead of relying on a single method, this algorithm probabilistically selects between Order Crossover and One-Point Crossover. This specific combination allow the solver to alternate between preserving job sequences and disrupting stagnant schedule structures, effectively helping the population escape local optima.

By Generation 40, the Meta-Controller shifts focus toward the quality of gene propagation by introducing Algorithm S99. This algorithm abandons random parent selection in favor of a Ranked Selection mechanism, which ensures that higher-performing parents are more likely to contribute to the next generation. To prevent premature convergence caused by this selection pressure, the algorithm increases the mutation rate to $0.25$ and utilizes adaptive multi-point adjustments. This trend of refining the selection logic culminates in Generation 55 with Algorithm S118, which implements a Diversity-Fitness Dual Selection. This two-layer approach filters parents based on their diversity contribution before applying fitness ranking. Furthermore, it enhances the local search by dynamically extending the search depth if consecutive improvements are detected, allowing the solver to adapt its intensity to the local landscape geometry.

\subsection{Domain Awareness and Final Convergence}A significant turning point in the trajectory occurs in Generation 95 with the synthesis of Algorithm S175. Landscape analysis at this stage reveals that generic genetic operators are no longer yielding significant gains. Consequently, DyACE generates a Compatibility-aware Heuristic Crossover designed to prioritize job-machine assignment compatibility. By shifting from simple permutations to a logic that understands the physical constraints of the Job Shop Scheduling Problem, the algorithm achieves more precise structural optimizations. To complement this domain-specific logic, the local search iterations are doubled to 100, focusing on intense neighborhood refinement to capitalize on high-quality solutions.

The evolution concludes with Algorithm S304 in Generation 140, which establishes a final refinement phase to secure global optimality. This algorithm employs a Variable-depth Dynamic Optimization logic that uses a feedback loop to adjust local search iterations based on previous improvement success. It also utilizes a hybrid crossover that dynamically toggles between single-point and two-point methods based on parental fitness. This highly adaptive configuration provides the precision necessary for the final convergence stage, successfully navigating the remaining complexity of the ta51 instance and reaching the final optimized schedule.

\section{Computational Cost Analysis}
\label{app:cost_analysis}

This section provides a detailed analysis of the computational overhead associated with the DyACE framework. Because perturbative heuristics rely on the iterative evolution of a population, evaluating each algorithmic logic requires multiple Monte Carlo simulations, which is inherently expensive. By reframing the static offline search as a dynamic online control problem, DyACE maintains a unified evaluation budget while significantly reducing the computational cost incurred during each specific evaluation step.

\subsection{Evaluation Efficiency via Short-horizon Rollout}

In traditional offline algorithm design workflows, such as those used in FunSearch or ReEvo, every candidate algorithm must be evaluated over a complete evolutionary cycle (e.g., 150 generations) to obtain a final performance metric. This process results in a heavy computational burden for each evaluation. In contrast, DyACE introduces the Look-Ahead Rollout Search mechanism. By performing local evaluations on the current population state over a shorter horizon of 30 generations, the system rapidly captures the ``short-horizon potential'' of an algorithm without waiting for the entire trajectory to conclude. Consequently, although DyACE requires re-evaluation at each decision step to account for the non-stationary landscape, the actual computational cost per evaluation is substantially lower than that of offline baselines.

\subsection{LLM Inference Overhead}

Regarding inference costs, frameworks like FunSearch and EoH require only a single LLM call to generate a new algorithm. DyACE's Meta-Controller introduces additional inference overhead, as the generation of each new algorithm involves a two-stage process: decoupled diagnosis and coding. Despite this increased complexity per generation, DyACE maintains an algorithm population size of $M=5$ and incorporates a retry mechanism for failed generations to ensure stability. Under the constraint of 300 total function evaluations, the cumulative LLM inference cost for DyACE remains comparable to that of EoH and FunSearch. Conversely, the ReEvo method also introduces extra inference steps through its ``reflection-generation'' loop. However, because ReEvo employs a larger algorithm population size of $M=10$, it often incurs double the total LLM inference overhead compared to other methods within the same evaluation budget.

\subsection{Scalability and Future Optimizations}

Empirical evidence suggests that the advantages of DyACE become increasingly significant as problem dimensions, such as the number of jobs or machines, expand. In large-scale instances, static algorithms often encounter a ``scalability wall'' where fixed logic fails to adapt to the complex, shifting landscape geometry of high-dimensional spaces. While DyACE introduces inference costs, its ability to adjust search operators in real-time reduces the total generations required to reach a target gap. Future research will explore a hierarchical meta-control structure that utilizes smaller, locally deployed models for frequent parameter fine-tuning, reserving large foundation models for major strategic algorithmic changes. This approach is expected to further reduce inference latency, allowing DyACE to be applied to real-time dispatching scenarios with strict timing requirements.

\end{document}